\documentclass{article}

\usepackage{amsmath}
\usepackage{amssymb}
\usepackage{times}
\usepackage{bm}
\usepackage{xargs}
\usepackage{multirow}
\usepackage{amsfonts}
\usepackage{color}
\usepackage{booktabs}
\DeclareMathOperator*{\argmin}{arg\,min}

\newcommandx\includeImageLineWidth[2][1=1.0]{\includegraphics[width=#1\linewidth]{#2}}

\newcommand{\bc}{{\bm{c}}}

\newcommand{\bs}{{\bm{s}}}

\newcommand{\bx}{{\bm{x}}}

\newcommand{\eg}{{\emph{g}}}

\usepackage{array}
\newcommand{\PreserveBackslash}[1]{\let\temp=\\#1\let\\=\temp}
\newcolumntype{C}[1]{>{\PreserveBackslash\centering}p{#1}}
\newcolumntype{R}[1]{>{\PreserveBackslash\raggedleft}p{#1}}
\newcolumntype{L}[1]{>{\PreserveBackslash\raggedright}p{#1}}

\def\eg{\emph{e.g.,}}           

     \PassOptionsToPackage{numbers, compress}{natbib}

\usepackage[preprint]{neurips_2024}

\usepackage[utf8]{inputenc} 
\usepackage[T1]{fontenc}    
\usepackage[colorlinks,citecolor=teal]{hyperref}       
\usepackage{url}            
\usepackage{booktabs}       
\usepackage{amsfonts}       
\usepackage{nicefrac}       
\usepackage{microtype}      
\usepackage{xcolor}         
\usepackage{algorithm}
\usepackage{algorithmic}
\usepackage{graphicx}
\usepackage{subfigure}
\usepackage{subfiles}
\usepackage{enumitem}
\usepackage{wrapfig}
\usepackage{capt-of}
\usepackage{pifont}

\usepackage{listings}

\usepackage{xcolor}

\definecolor{codegreen}{rgb}{0,0.6,0}
\definecolor{codegray}{rgb}{0.5,0.5,0.5}
\definecolor{codepurple}{rgb}{0.58,0,0.82}
\definecolor{backcolour}{rgb}{0.95,0.95,0.92}

\lstdefinestyle{mystyle}{
  backgroundcolor=\color{backcolour}, commentstyle=\color{codegreen},
  keywordstyle=\color{magenta},
  numberstyle=\tiny\color{codegray},
  stringstyle=\color{codepurple},
  basicstyle=\ttfamily\footnotesize,
  breakatwhitespace=false,         
  breaklines=true,                 
  captionpos=b,                    
  keepspaces=true,                 
  numbers=left,                    
  numbersep=5pt,                  
  showspaces=false,                
  showstringspaces=false,
  showtabs=false,                  
  tabsize=2
}

\title{Generative Lifting of Multiview to 3D from Unknown Pose: Wrapping NeRF inside Diffusion}

\author{%
Xin Yuan\\%
University of Chicago%
\And
Rana Hanocka\\%
University of Chicago%
\And
Michael Maire\\%
University of Chicago%
\AND%
~\vspace{-2.75em}\\%
\normalfont{\texttt{\{yuanx,ranahanocka,mmaire\}@uchicago.edu}}\\%
}

\newcommand{\xmark}{\ding{55}}%

\begin{document}

\maketitle

\begin{abstract}
We cast multiview reconstruction from unknown pose as a generative modeling problem.  From a collection of unannotated 2D images of a scene, our approach simultaneously learns both a network to predict camera pose from 2D image input, as well as the parameters of a Neural Radiance Field (NeRF) for the 3D scene.  To drive learning, we wrap both the pose prediction network and NeRF inside a Denoising Diffusion Probabilistic Model (DDPM) and train the system via the standard denoising objective.  Our framework requires the system accomplish the task of denoising an input 2D image by predicting its pose and rendering the NeRF from that pose.  Learning to denoise thus forces the system to concurrently learn the underlying 3D NeRF representation and a mapping from images to camera extrinsic parameters.  To facilitate the latter, we design a custom network architecture to represent pose as a distribution, granting implicit capacity for discovering view correspondences when trained end-to-end for denoising alone.  This technique allows our system to successfully build NeRFs, without pose knowledge, for challenging scenes where competing methods fail.  At the conclusion of training, our learned NeRF can be extracted and used as a 3D scene model; our full system can be used to sample novel camera poses and generate novel-view images.

\end{abstract}

\section{Introduction}

Structure from motion is a well-studied problem in computer vision, with a substantial history of research focusing on
the specific task of reconstructing a 3D scene from a collection of 2D images captured from different viewpoints.
When the 3D pose (camera extrinsics) for each 2D view is unknown, classic approaches~\cite{snavely2006photo,
agarwal2011building} explicitly estimate correspondence between 2D views (\eg~by matching local feature descriptors)
prior to optimizing a shared 3D geometry whose reprojections are consistent with those views.  Neural Radiance Fields
(NeRFs)~\cite{DBLP:conf/eccv/MildenhallSTBRN20,DBLP:journals/corr/abs-2111-12077,DBLP:conf/cvpr/Martin-BruallaR21,
DBLP:conf/iccv/BarronMVSH23} have led a revolution toward widespread use of differentiable 3D scene
representations~\cite{DBLP:conf/eccv/MildenhallSTBRN20,fridovich2022plenoxels,kerbl20233d} that are compatible with deep
learning techniques.  However, the problem of jointly solving for both the 3D reconstruction and the pose, when neither is
known a priori, remains an open problem. Recent attempts to connect learning of camera pose with NeRFs operate under
simplifying assumptions, such as coarse pose initialization (only learning adjustments)~\cite{lin2021barf} or
front-facing (as opposed to arbitrary 360$^{\circ}$) views of the scene~\cite{wang2021nerfmm}.

\begin{figure*}[tbh]
   \begin{center}
      \includegraphics[width=\linewidth]{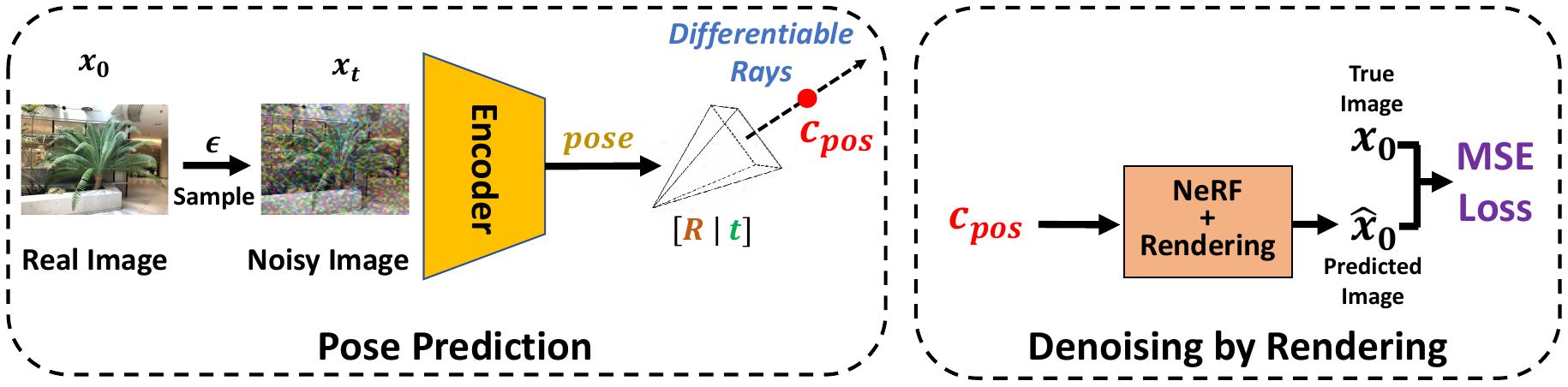}
   \end{center}
   \vspace{-0.75em}
   \caption{%
      \textbf{Wrapping NeRF inside Diffusion.}
      We learn a 3D scene reconstruction by training a denoising diffusion model (DDPM) on a dataset of 2D views of
      the scene.  The architecture of our DDPM consists of two components.
      \emph{\textbf{Left:}}
	 An \emph{Encoder} predicts the pose of a single noisy 2D input image.
      \emph{\textbf{Right:}}
	 A \emph{NeRF} is rendered from the predicted camera pose to create a 2D output image that is treated as the
         predicted denoising of the input view.
      The system must learn parameters of both the \emph{Encoder} and \emph{NeRF} so that any 2D view can be denoised
      by predicting a camera and rendering the scene.  The NeRF rendering process is differentiable with respect to
      rays shot from the camera, which themselves depend on the camera-to-world transformation matrix produced by the
      encoder.  All modules are end-to-end trainable, and the system is optimized by the simple MSE loss on denoising.%
   }%
   \label{fig:framework}
\end{figure*}

In parallel with the development of differentiable 3D representations, progress across a variety of paradigms for
generative models~\cite{goodfellow2014generative,DBLP:journals/corr/KingmaW13,DBLP:conf/nips/HoJA20}, has
transformed the landscape for designing and training systems using deep learning.  Learning to synthesize data provides
an unsupervised training objective and scaling compute, parameters, and datasets is a path toward foundation
models~\cite{bommasani2021opportunities} whose feature representations can subsequently be repurposed to specific
downstream tasks.  However, large-scale foundation models are not the only setting in which generative learning is
appropriate.  Nor is manipulation of pre-trained models (\eg~extracting features, fine-tuning, or prompting) the
only strategy for applying generative learning to solve downstream tasks.

\citet{yuan2023factorized} demonstrate an alternative strategy that utilizes a generative model and relies solely on a
generative learning objective, yet directly solves a downstream task (image segmentation) as a byproduct
of training the generative model.  Their strategy is to constrain the architecture of the generative model such that it
must synthesize an image by first predicting a segmentation and then generating the corresponding image regions in
parallel.  Trained as a Denoising Diffusion Probabilistic Model (DDPM)~\cite{DBLP:conf/nips/HoJA20}, segmentation emerges
as the bottleneck representation in a network that first partitions a noisy input into regions and then denoises each
region in parallel.

We port this general concept to the problem of multiview 3D reconstruction from unknown pose, where we devise an internal
pose prediction network and a NeRF comprising the task-specific architecture encapsulated within our DDPM; see
Figure~\ref{fig:framework}.  We solve a small-scale generative modeling problem: learning to generate images in the
collection of 2D views of a single scene.  Training examples are noised 2D images (views), the DDPM output is a predicted
denoised image, and the loss is the denoising objective.  Inside our generative wrapper, the model
architecture dictates that we map a noisy image to a predicted camera pose and then render the NeRF from that pose to
synthesize the clean output image.  Successfully performing denoising in this manner requires that: (a) the NeRF
stores a 3D scene representation consistent with all of the 2D views, and (b) the pose prediction network implicitly
solves the 2D view correspondence problem by mapping each 2D input image to camera coordinates from which it is
reconstructed by rendering the NeRF.

\begin{figure*}[tbh]
   \begin{center}
      \includegraphics[width=\columnwidth]{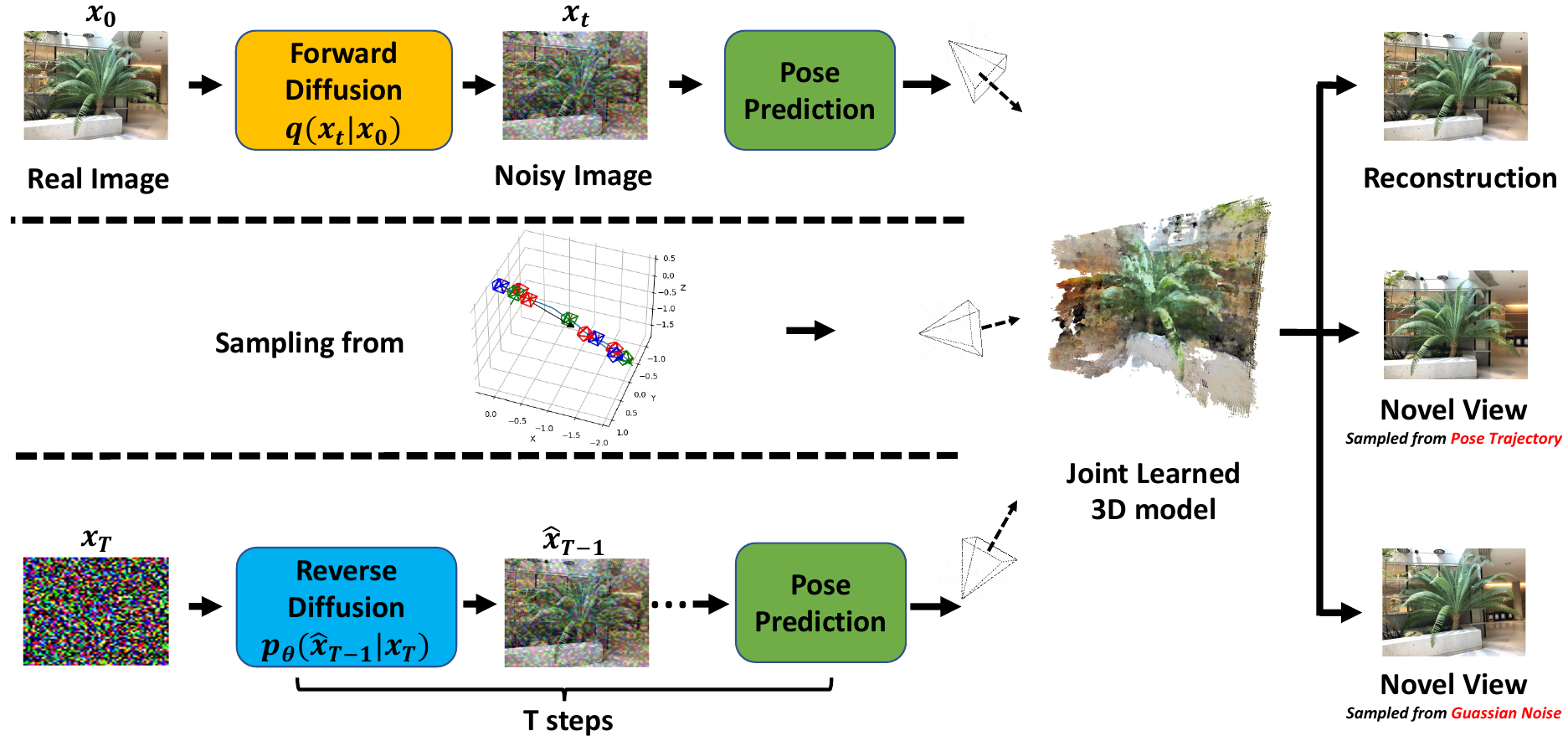}
   \end{center}
   \vspace{-0.75em}
   \caption{%
      \textbf{Unifying pose prediction, 3D reconstruction, and novel-view image generation.}
      Our trained system (Figure~\ref{fig:framework}) can be deployed for multiple tasks.
      \emph{\textbf{Pose prediction (top):}}
         We can predict the pose of a previously unseen real image by adding a small amount of noise (forward
         diffusion) and feeding it to our \emph{Encoder} (Fig~\ref{fig:framework}, \emph{left}).  Rendering our
         learned \emph{NeRF} from that camera pose should reconstruct the real image.
      \emph{\textbf{Direct NeRF usage (middle):}}
         Our learned \emph{NeRF} can be extracted and directly used to render the scene (\eg~along a manually
         specified camera path).
      \emph{\textbf{Sampling cameras and views (bottom):}}
         Performing sequential diffusion denoising from pure Gaussian noise input synthesizes a camera pose
         from which rendering the \emph{NeRF} generates a novel view of the scene.%
   }%
   \label{fig:overview}
\end{figure*}

\begin{figure*}[tbh]
   \begin{center}
      \includegraphics[width=\linewidth]{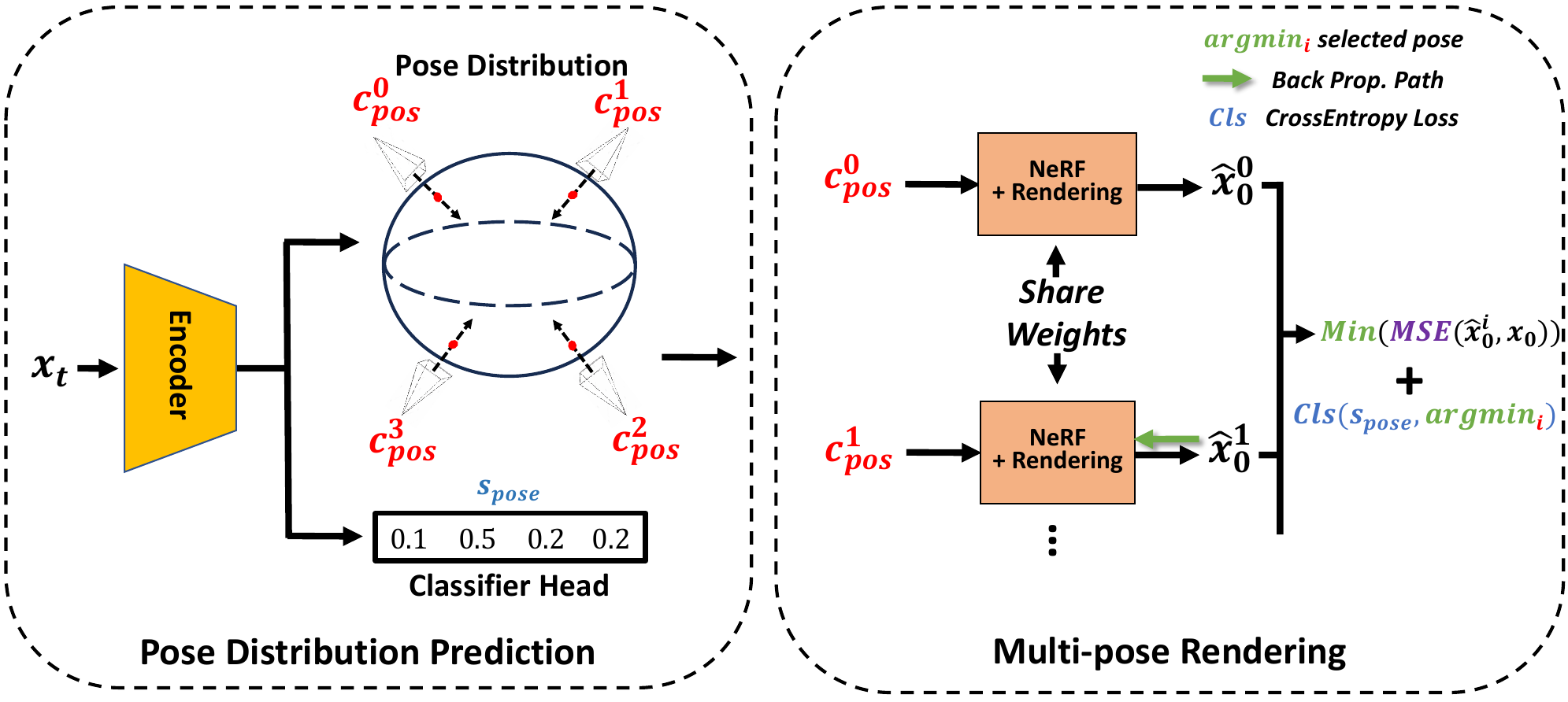}
   \end{center}
   \vspace{-0.75em}
   \caption{%
      \textbf{Pose distribution representation and multi-pose rendering for 360$^{\circ}$ scenes.}
      In order to perform view denoising by learning a NeRF and predicting the pose from which to render it, our
      system (Figure~\ref{fig:framework}) must implicitly solve multiview correspondence by mapping training images
      (of unknown pose) into consistent locations in the 3D environment.  We enable training via gradient descent
      to discover such solutions for challenging multiview datasets (\eg~spanning 360$^{\circ}$) by augmenting our
      architecture with the capacity to represent uncertainty over a pose distribution.
      \emph{\textbf{Left:}}
         Our encoder, given a noisy image $\bx_t$, predicts parameters for multiple cameras and a corresponding 
         probability distribution over cameras, $\bs_{pose}$.
      \emph{\textbf{Right:}}
         During training, we render the NeRF from each predicted camera and use the best reconstruction to 
         calculate the denoising loss; an auxiliarly classification loss pushes the predicted camera distribution
         to upweight the selected output.
      At test time, we render using only the single camera predicted as most likely by the classifier.%
   }%
   \label{fig:factpose}
\end{figure*}

Figure~\ref{fig:overview} illustrates how our trained system jointly solves pose prediction and 3D reconstruction.
Our system enables predicting the 3D pose for new (unseen) images, and re-rendering the learned scene from different
camera poses which can be generated from noise or provided explicitly.  As Figure~\ref{fig:factpose} shows and
Section~\ref{sec:method} describes in detail, we significantly expand the complexity of multiview reconstruction
problems our system can solve by replacing our simple pose prediction network with a more expressive version.  This
alternative maintains a representation of uncertainty over a distribution of multiple possible poses, which gives our
system, trained from scratch by gradient descent, the implicit capacity to explore more view correspondence
configurations.

Our contributions are:
\vspace{-0.75em}
\begin{itemize}[leftmargin=0.2in]
   \item{%
      A new approach to 3D reconstruction from unknown pose based entirely on generative training.  Denoising
      is a generative wrapper that ``lifts'' an architecture consisting of a forward model for pose prediction and
      differentiable rendering to learn view correspondence and 3D reconstruction.  Compared to an autoencoder, this
      fully generative wrapper benefits learned reconstruction quality.%
   }%
   \vspace{-0.25em}
   \item{%
      A novel architecture for pose prediction that enables representing uncertainty during training,
      allowing us to learn 3D reconstructions from 2D views of arbitrary and unknown pose.
   }%
   \vspace{-0.25em}
   \item{%
      New capabilities for 3D NeRF reconstruction which are demonstrated through experiments on arbitrary 360-degree
      poses.  While \citet{wang2021nerfmm} can reconstruct under certain assumptions about an unknown camera
      (\eg~forward-facing views of the scene), they fail on image collections from unconstrained pose
      (\eg~360$^{\circ}$ views).  Our method successfully reconstructs a NeRF and infers camera pose for these
      challenging datasets.%
   }%
\end{itemize}

\section{Related Work}
\label{sec:related}

\noindent \textbf{Neural Radiance Fields (NeRFs)}~\cite{DBLP:conf/eccv/MildenhallSTBRN20} have emerged as a powerful
framework for 3D scene reconstruction and view synthesis, with multiple extensions and improvements~\cite{
DBLP:journals/corr/abs-2111-12077,DBLP:conf/cvpr/Martin-BruallaR21,DBLP:conf/iccv/BarronMVSH23}.
NeRF${+}{+}$~\cite{zhang2020nerf++} adds spatially-varying reflectance and auxiliary tasks for better training.
PixelNeRF~\cite{yu2021pixelnerf} extends NeRF to generate high-quality novel views from one or few input
images, but still requires camera pose information.
NeRF${-}{-}$~\cite{wang2021nerfmm} jointly optimizes camera intrinsics and extrinsics as learnable parameters while
training NeRFs.  However, their proposed training scheme and camera parameterization cannot handle large camera
rotation and is restricted to forward-facing views of the scene.

\noindent \textbf{Generative models for 3D reconstruction} aim to infer the underlying 3D structure of a scene from a
set of 2D images.  These models often learn a latent representation of the 3D scene and use it to generate novel views
or perform other tasks such as object manipulation or scene editing.  Generative Radiance Fields
(GRAFs)~\cite{schwarz2020graf} combine NeRFs with VAEs or GANs to generate novel 3D scenes without explicit 3D
geometry.  GRAFs learn a latent space encoding scene structure, with NeRF mapping points in this space to 3D radiance
fields.  DiffRF~\cite{muller2023diffrf} leverages the diffusion prior to perform 3D completion in a two-stage manner,
which is further improved by SSD-NeRF~\cite{ssdnerf} with a single stage training scheme and an end-to-end objective
that jointly optimizes a NeRF and diffusion.  Multiple works combine NeRF with generative models for the purpose of
3D synthesis~\cite{chan2021pi,meng2021gnerf,gu2021stylenerf}, including ones that place NeRF and diffusion models
in series~\cite{poole2022dreamfusion,lin2023magic3d,wang2023score}.  Our framework's nested structure differs, as
our aim is not to learn to generate novel 3D scenes; we aim to use generative training to solve the classic
multiview 3D reconstruction problem.

\noindent \textbf{Pose estimation} is the challenging task of estimating object or camera position and orientation
within a scene.  COLMAP~\cite{schoenberger2016sfm,schoenberger2016mvs} uses a Structure-from-Motion (SfM)~\cite{
schoenberger2016sfm} approach for pose estimation, can handle challenging scenes with varying lighting conditions and
viewpoints, and is widely used in NeRF training.  However, this collection of techniques requires a large number
of images for accurate pose estimation; pre-processing also restricts flexibility.  PoseDiffusion~\cite{
wang2023posediffusion} and Camera-as-Rays~\cite{zhang2024cameras} use a diffusion model to denoise camera parameters
and rays.  Although sharing similar spirit in adopting diffusion, these methods require a supervised pertaining stage.
More importantly, diffusion serves as a different role in our model: instead of denoising cameras to recover the pose
distribution, we modulate a pose prediction system embedded inside the diffusion training process, yielding pose
information as a latent representation.

\section{Method}
\label{sec:method}


\subsection{Unsupervised Pose Prediction from a Single Image}
Our pose module (Figure~\ref{fig:framework}, left) consists of several components designed to predict, from a 2D image,
the position and orientation of a camera in the scene.  We design the encoder based on a standard DDPM U-Net.
We obtain input $\bx_t$, a noise version of $\bx_0$, via forward diffusion:
\begin{eqnarray} \label{eq:f_diff}
   q(\bx_t|\bx_0):= \mathcal{N}(\bx_t; \sqrt{\bar{\alpha}_t}\bx_0, (1-\bar{\alpha}_t)I), \nonumber \\
   \bx_{t} = \sqrt{\bar{\alpha}_t}\bx_0 + \sqrt{1-\bar{\alpha}_t}\epsilon, \epsilon \sim \mathcal{N}(0,1),
\end{eqnarray}
where $\alpha_t = 1 - \beta_t, \bar{\alpha}_t = \prod_{s=1}^t \alpha_t$.
We encode the pose information in the form of a camera-to-world transformation matrix, \(T_{wc} = [Ro|ts]\),
where \(Ro \in SO(3)\) represents the camera's rotation, and \(ts \in \mathbb{R}^3\) represents its translation.
Following~\cite{wang2021nerfmm}, we adopt Rodrigues' formula to form the rotation matrix \(Ro\) from axis-angle
representation:
\begin{equation}\label{eq:Ro}
    Ro = I + \sin(\phi) [\omega]_{\times} + (1 - \cos(\phi)) [\omega]_{\times}^2
\end{equation}
where \(\phi\) is the rotation angle, \(\omega\) is a normalized rotation axis, and \([\omega]_{\times}\) is the
skew-symmetric matrix of the rotation axis vector \(\omega\).

Different from~\cite{wang2021nerfmm}, which formulates \(\omega_i\) and translation \({ts}_i\) as trainable
parameters, for each input image \(\bx^i\), our system maps the corresponding U-Net encoder features to two
$3$-dimensional input-dependent feature vectors.  By doing so, we not only learn to fit the pose information during
training, but also obtain a pose predictor network applicable to any input 2D image.

\subsection{3D Optimization with Denoising Rendering}
With the predicted camera pose from the U-Net Encoder, the system is able to perform denoising via differentiable
rendering.  Specifically, as shown in the right side of Figure~\ref{fig:framework}, we sample the differentiable
coordinates $\bc_{pos}$, which are fed into a NeRF MLP model to learn object density and opacity, generating a
2D image reconstruction $\hat{\bx}_0$.

Benefiting from the compatibility between NeRF reconstruction loss and DDPM denoising objective, we train our model
end-to-end by simply minimizing the pixel-wise distance from $\hat{\bx}_0$ to ${\bx}_0$.  Model weights of the camera
predictor (U-Net encoder) and NeRF MLPs are optimized with loss:
\begin{eqnarray} \label{eq:loss1}
   L = \mathbb{E}||\hat{\bx}_0 - {\bx}_0||_2^2
\end{eqnarray}
Algorithm~\ref{alg:training1} summarizes training.

\begin{figure}[tp]
\begin{minipage}[t]{0.412\columnwidth}
   \begin{algorithm}[H]
   \footnotesize
   \caption{Generative Lifting to 3D:\\Single Camera Pose Prediction \& NeRF}
   \label{alg:training1}
   \begin{algorithmic}
      \STATE {\bfseries Input:} Multiview image collection $\bm{\mathcal{X}}$
      \STATE {\bfseries Output:} Encoder (pose predictor) \& NeRF
      \STATE {\bfseries Initialize:} Model weights $\theta$, Timesteps $T$
      \FOR{$\text{iter}=1$ {\bfseries to }Iter$_{total}$}
         \STATE Sample $\bm{x}_0 \in \bm{\mathcal{X}}$, $t \in [1,T]$.
         \STATE Sample $\bx_t$ using Eq.~\ref{eq:f_diff}.
         \STATE Predict $Ro$ using Eq.~\ref{eq:Ro} and $ts$.
         \STATE Compute $\hat{\bx}_0$ by rendering the NeRF from the pose predicted for $\bx_t$
                (see Figure~\ref{fig:framework}).
         \STATE Backprop from loss in Eq.~\ref{eq:loss1}.
         \STATE Update model weights.
     \ENDFOR
     \STATE return $\theta$
   \end{algorithmic}
   \end{algorithm}
\end{minipage}
\hfill
\begin{minipage}[t]{0.575\columnwidth}
   \begin{algorithm}[H]
   \footnotesize
   \caption{Generative Lifting to 3D using Pose Distribution Modeling \& Multi-pose NeRF Rendering}
   \label{alg:training2}
   \begin{algorithmic}
      \STATE {\bfseries Input:} Multiview image collection $\bm{\mathcal{X}}$
      \STATE {\bfseries Output:} Encoder (multi-pose predictor \& classifier) \& NeRF
      \STATE {\bfseries Initialize:} Model weights $\theta$, Timesteps $T$, and initial pose of candidate cameras
      \FOR{$\text{iter}=1$ {\bfseries to }Iter$_{total}$}
         \STATE Sample $\bm{x}_0 \in \bm{\mathcal{X}}$, $t \in [1,T]$.
         \STATE Sample $\bx_t$ using Eq.~\ref{eq:f_diff}.
         \STATE Predict poses and corresponding $\bs_{pose}$, as in Figure~\ref{fig:factpose}.
         \STATE Compute $\{\hat{\bx}_0^i\}$ using multi-pose rendering (Figure~\ref{fig:factpose}).
         \STATE Backprop along the path of the best render via Eq.~\ref{eq:loss2}.
         \STATE Update model weights.
     \ENDFOR
     \STATE return $\theta$
   \end{algorithmic}
   \end{algorithm}
\end{minipage}
\end{figure}

\subsection{Multi-pose Rendering for Scene Reconstruction from 360$^{\circ}$ Views}
A failure to estimate poses accurately can occur when rotation perturbations exceed a certain threshold in
Eq.~\ref{eq:Ro}, which prevents learning from 360$^{\circ}$ views of a scene~\cite{wang2021nerfmm}.  Even with good
reconstruction in 2D space, an overfitting issue can occur during optimization, where NeRF compensates by creating
multiple disjoint copies of scene fragments instead of a unified 3D reconstruction.  We address these issues via a
higher capacity pose predictor capable of representing uncertainty (Figure~\ref{fig:factpose}).

\noindent \textbf{Pose distribution prediction.}
A simple camera parameterization is to restrict position to a fixed-radius sphere with fixed intrinsics, and the
constraint of always looking towards the origin at (0, 0, 0).  Assuming the object in a 360$^{\circ}$ scene is
centered, rotated, and scaled by some canonical alignment, such a parametrization has only two degrees of freedom.
However, this simplistic approach restricts model capacity for capturing diverse viewpoints or extensive rotations.

We propose a more flexible approach that allows for a wider range of camera positions and orientations.  Given a 2D
image captured from a specific viewpoint, instead of predicting a single transformation, we sample the camera's
position from a distribution of multiple cameras that cover a larger range of positions and orientations on a sphere.

As Figure~\ref{fig:factpose} shows, different candidate cameras spread over the sphere, pointing to the origin at
initialization.  Each input view predicts parameters for all cameras in the distribution.  An auxiliary classifier
head predicts the probability of the input view corresponding to each camera in the distribution.  Early in training,
such a design facilitates searching over multiple pose hypotheses in order to discover a registration of all views
into a consistent coordinate frame.  Only one camera prediction per input view need be correct, as long as the system
also learns which one.

\noindent \textbf{Joint optimization with multi-pose rendering.}
We render the NeRF separately from each candidate camera to produce a set of 2D images $\{\hat{\bx}_0^i\}$.  During
backpropagation, we only allow the gradient to pass selectively to optimize the best match between the true image and
the rendered reconstruction.  The selected branch index serves as a pseudo-label to co-adapt the classifier head in a
self-supervised bootstrapping manner.  The total loss for joint training is:
\begin{eqnarray}\label{eq:loss2}
L = \min_{i} ||\hat{\bx}_0^i - {\bx}_0||_2^2 + \lambda CrossEntropy(\bs_{pose}, \argmin_{i}||\hat{\bx}_0^i - {\bx}_0||_2^2)
\end{eqnarray}
where $\lambda$ is the trade-off parameter between view reconstruction and camera classification.  We set
$\lambda$ as 0.1 in experiments and investigate its effect in an ablation study.  Algorithm~\ref{alg:training2}
summarizes training.

\subsection{Novel View Generation}
Figure~\ref{fig:overview} illustrates the different modalities in which our trained system can be used.

\noindent \textbf{Pose prediction \& reconstruction.}
Given input image $\bx_0$, we sample a noisy version $\bx_t$ through forward diffusion in
Eq.~\ref{eq:f_diff}.  We then pass $\bx_t$ to the model.  Our system estimates the camera pose of $\bx_t$ with
respect to the scene and recovers a clean image reconstruction with one-step denoising.

\noindent \textbf{Sampling from pose trajectory}. Though not trained with any camera pose information, our system can
generate novel views using a pre-defined camera trajectory, acting as a conventional NeRF model.

\noindent\textbf{Sampling from Gaussian noise.}
A unique property of our system is its support for sampling cameras and scene views.  Using reverse diffusion, our
model can generate a realistic novel view and the corresponding camera pose, starting from a pure noise input
$\bx_T \sim \mathcal{N}(0,1)$.  We perform $T$ steps of reverse diffusion (predict $\bx_{t-1}$ from $\bx_{t}$) to
progressively generate a novel view.

\section{Experiments}

We conduct the experiments on the face-forwarding dataset LLFF~\cite{mildenhall2019llff} with a resolution of
$378 \times 504$, using the single camera prediction system depicted in Alg.~\ref{alg:training1}.  To handle
$360^\circ$ scenes, we adopt the multi-pose rendering as in Alg.~\ref{alg:training2} on ShapeNet
Car~\cite{DBLP:journals/corr/ChangFGHHLSSSSX15} (5 scenes), Lego and Drums~\cite{DBLP:conf/eccv/MildenhallSTBRN20}
with a resolution of $128 \times 128$.  Instead of simultaneously optimizing two networks: one `coarse' and one `fine',
we only use a single network to represent $360^\circ$ scenes for all methods.  We initialize 8 camera candidates
spread over 8 quadrants of a sphere for the ShapeNet Car scene, and 12 candidates over 4 quadrants of a semi-sphere
for Lego and Drums.

\subsection{Implementation Details}
For all experiments, we use a U-Net~\cite{DBLP:conf/miccai/RonnebergerFB15} encoder as the pose prediction
module.
The downsampling stack performs five steps of downsampling, each with 2 residual blocks.
From highest to lowest resolution, U-Net stages use $[C,C,2C,2C,4C]$ channels, respectively.
We set $C = 64$ for all models.  Figure~\ref{fig:3darch_detail} details the network architecture.
We use 100 denoising steps for all models.

Our method involves two sets of trainable parameters: NeRF model weights and pose prediction network weights; we adopt
separate Adam optimizers, with learning rates $1e^{-4}$ and $2e^{-5}$ for NeRF and pose prediction, respectively.  We
set $(\beta_1,\beta_2)$ as $(0.9, 0.999)$ for both optimizers.  We adopt the same batch size and learning rate
scheduler used to train the corresponding baseline NeRF as in~\cite{DBLP:conf/eccv/MildenhallSTBRN20}.  We train all
models for 200k iterations.  Code segments~\ref{lst:transform_xyz},~\ref{lst:lkat} and~\ref{lst:pose_predict} detail
our camera parameterization.

\begin{figure*}[t]
   \begin{center}
   \begin{minipage}[t]{\linewidth}
      \vspace{0pt}
      \centering
      \subfigure[\textbf{\textsf{\scriptsize{Ground Truth 2D View}}}]{
        \includegraphics[width=0.95\linewidth]{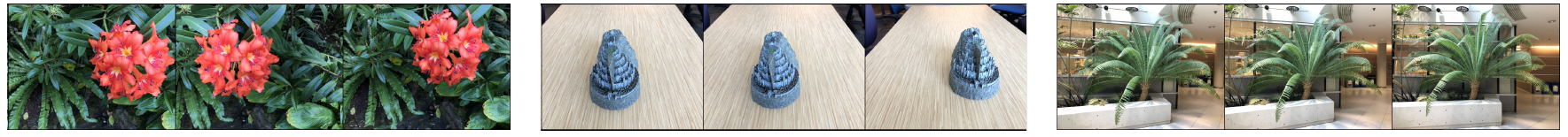}
      }
   \end{minipage}%
   \hfill%
   \begin{minipage}[t]{\linewidth}
      \vspace{0pt}
      \centering
      \subfigure[\textbf{\textsf{\scriptsize{2D Reconstruction}}}]{
         \includegraphics[width=0.95\linewidth]{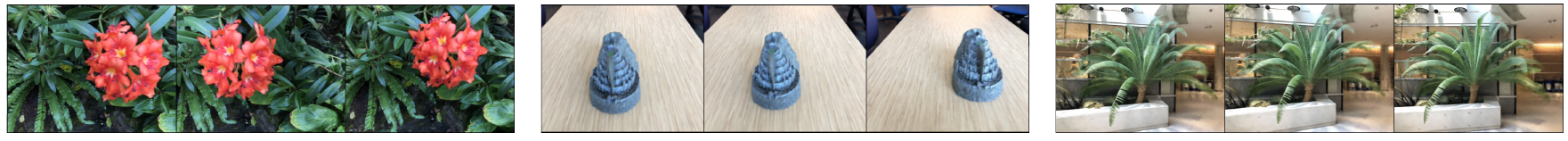}
      }
   \end{minipage}
   \hfill%
   \begin{minipage}[t]{\linewidth}
      \vspace{0pt}
      \centering
      \subfigure[\textbf{\textsf{\scriptsize{Predicted Disparity}}}]{
         \includegraphics[width=0.95\linewidth]{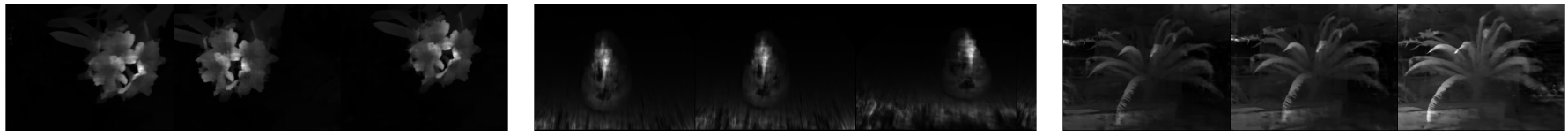}
      }
   \end{minipage}
    \hfill%
   \begin{minipage}[t]{\linewidth}
      \vspace{0pt}
      \centering
      \subfigure[\textbf{\textsf{\scriptsize{Point Cloud from NeRF}}}]{
         \includegraphics[width=0.95\linewidth]{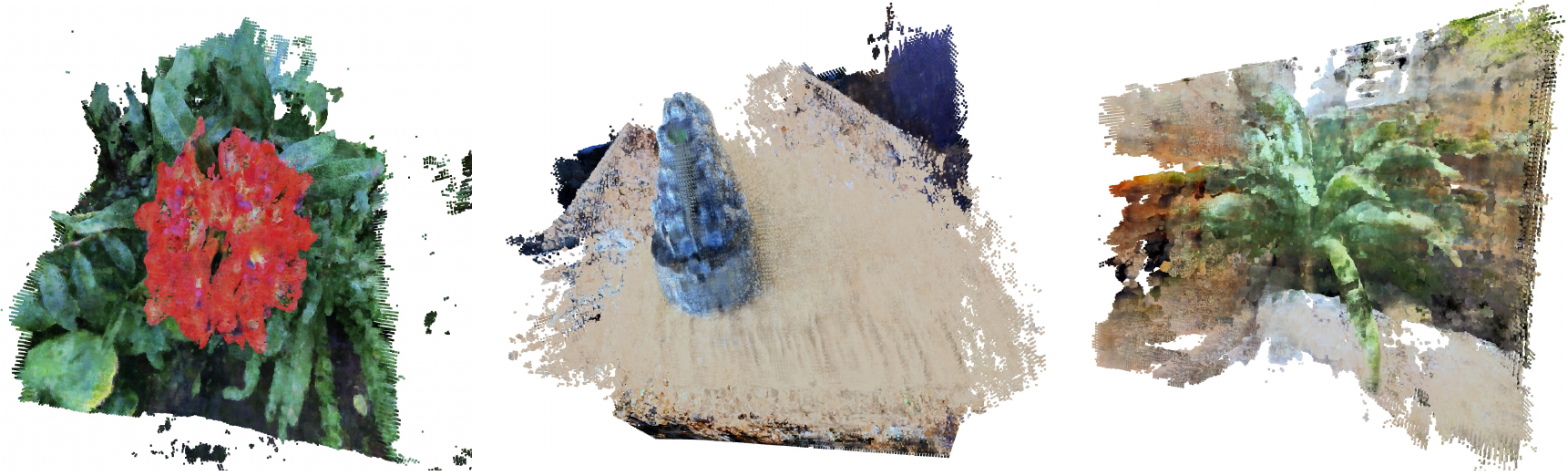}
      }
   \end{minipage}
   \end{center}
   \vspace{-1.25em}
   \caption{%
      \textbf{Reconstructions of images unseen during training on three scenes from LLFF~\cite{mildenhall2019llff}.}%
   }%
   \label{fig:llff_recon}
   \vspace{-1.0em}
\end{figure*}

\begin{table*}[t]
\footnotesize
\caption{%
   \textbf{Multiview reconstruction quality (PSNR, SSIM, \& LPIPS).}
   Our system, without pose knowledge, reconstructs 3D scenes from challenging image collections (views spanning
   $360^{\circ}$) on which NeRF{-}{-}~\cite{wang2021nerfmm} fails.  Supervised denotes standard NeRF training using
   ground-truth camera pose.%
}%
\vspace{0.25em}
\setlength{\tabcolsep}{2.9pt}
\label{tab:quant:2d}
\centering
\begin{tabular}{@{}llccccccccccccccc@{}}
\toprule
\multicolumn{1}{c}{} &&&
   \multicolumn{3}{c}{PSNR ($\uparrow$)} &&
   \multicolumn{3}{c}{SSIM ($\uparrow$)} &&
   \multicolumn{3}{c}{LPIPS ($\downarrow$)}\\
   [-2pt] \cmidrule{4-6} \cmidrule{8-10} \cmidrule{12-14}
\multicolumn{1}{@{}l}{\multirow{1}{*}{Type}} & \multicolumn{1}{l}{\multirow{1}{*}{Scene}} &&
   Supervised & NeRF{-}{-} & Ours &&
   \multicolumn{1}{c}{Supervised} & NeRF{-}{-} & Ours &&
   \multicolumn{1}{c}{Supervised} & NeRF{-}{-} & Ours\\[-1pt]
\midrule
& Fern   &     &     22.22 &21.67 &17.02   &     &0.64  &0.61  &0.42  & &0.47 &0.50 &0.55     \\
& Flower &     &     25.25 &25.34   &22.42   &  &0.71 &0.71   &0.58 & &0.36 &0.37 &0.43 \\
& Fortress  &        &  27.60 &26.20    &22.02    &     &0.73   &0.63 &0.50 &&0.38 &0.49 &0.51  \\
Forward- & Horns      &   &  24.25 &22.53  &17.48   &     &0.68 &0.61 &0.43  && 0.44 &0.50  &0.55\\ 
Facing & Leaves      &   &  18.81 &18.88  &14.44   &     &0.52   &0.53 &0.42 &&0.47 &0.47 &0.60 \\
& Orchids      &   &  19.09 &16.73 &14.34   &     &0.51  &0.39 &0.40 &&0.46 &0.55 &0.58 \\
&Room      &   &  27.77 &25.84 &22.36   &     &0.87  &0.84 &0.48 &&0.40 &0.44 &0.49 \\
& Trex      &   &  23.19 &22.67 &19.96   &     &0.74  &0.72 &0.62 &&0.41 &0.44 &0.51\\
\midrule
& Car && 28.98 &\xmark & 26.43 &&0.95 &\xmark &0.92 &&0.08 &\xmark &0.08\\
$360^{\circ}$ &Lego && 25.44 &\xmark &21.38 &&0.92 &\xmark &0.86 &&0.09 &\xmark &0.12 \\
& Drums &&22.12 &\xmark &18.65 &&0.89&\xmark &0.82  &&0.08 &\xmark &0.16 \\
\bottomrule
\end{tabular}
\end{table*}

\subsection{Multi-view 3D Reconstruction}
We measure the quality of reconstructions obtained by the \textit{top} pipeline in Figure~\ref{fig:overview}.
We directly input previously unseen images from different views to generate reconstructions.

\textbf{LLFF dataset.}
Figure~\ref{fig:llff_recon} shows we obtain good-quality reconstructions and disparity predictions.
Point cloud visualization plots the density and opacity output from our NeRF model at 3D coordinates.

\begin{figure*}[t]
   \begin{center}
   \begin{minipage}[t]{0.49\linewidth}
   \begin{minipage}[t]{0.25\linewidth}
      \vspace{0pt}
      \centering
      \subfigure[\textbf{\textsf{\scriptsize{Point Cloud.}}}]{
        \includegraphics[width=\linewidth]{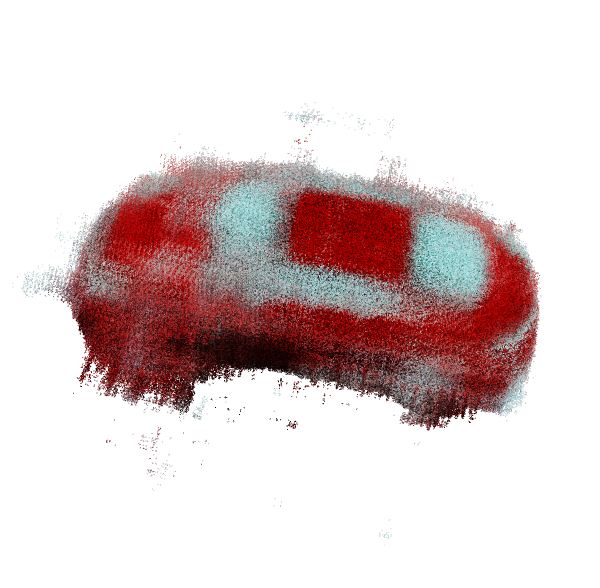}
      }
   \end{minipage}%
      \hfill
   \begin{minipage}[t]{0.7\linewidth}
      \vspace{0pt}
      \centering
      \subfigure[\textbf{\textsf{\scriptsize{2D reconstruction}}}]{
         \includegraphics[width=\linewidth]{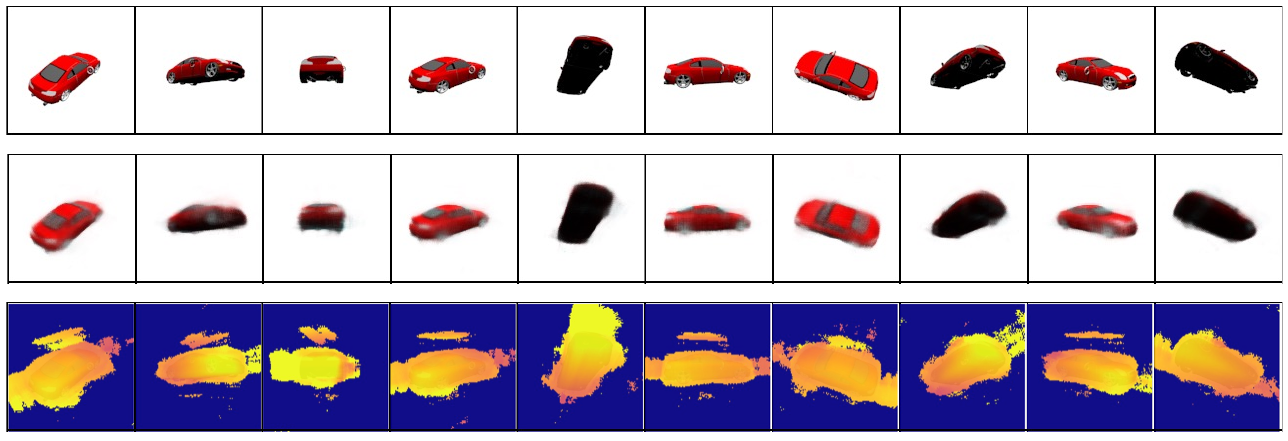}
      }
   \end{minipage}
   \end{minipage}
   \begin{minipage}[t]{0.49\linewidth}
   \begin{minipage}[t]{0.25\linewidth}
      \vspace{0pt}
      \centering
      \subfigure[\textbf{\textsf{\scriptsize{Point Cloud.}}}]{
        \includegraphics[width=\linewidth]{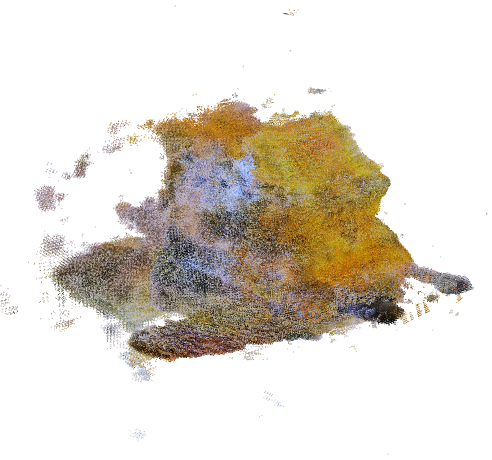}
      }
   \end{minipage}%
      \hfill
   \begin{minipage}[t]{0.7\linewidth}
      \vspace{0pt}
      \centering
      \subfigure[\textbf{\textsf{\scriptsize{2D reconstruction}}}]{
         \includegraphics[width=\linewidth]{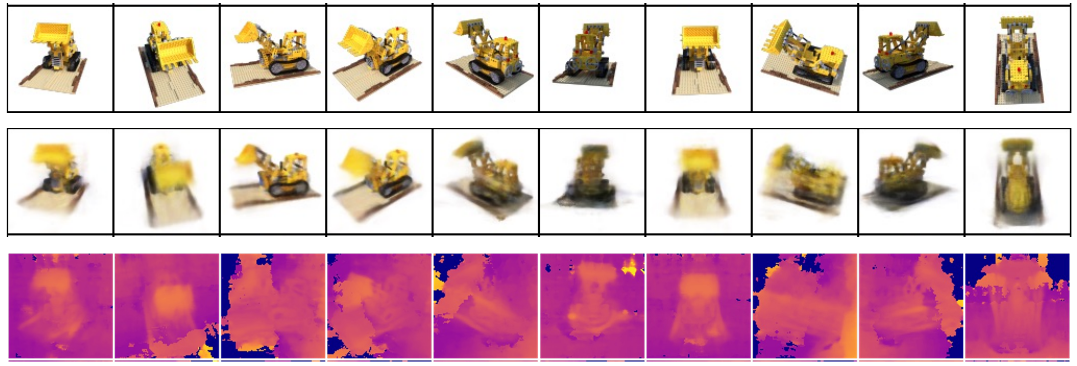}
      }
   \end{minipage}
   \end{minipage}
   \end{center}%
   \vspace{-1em}
   \caption{%
      \textbf{Reconstructions on $360^\circ$ scenes.}
   }%
   \label{fig:car_2d}
\end{figure*}

\textbf{$360^{\circ}$ scenes.}
Camera motions with large rotation perturbations cause failures in NeRF{-}{-};  it cannot handle $360^{\circ}$
scenes like Lego.  Our method solves this challenging case from a single input image, without relative pose estimation
between image pairs.  To obtain reconstructions, we determine the camera pose for the input image based on the maximum
score of the classification head in Figure~\ref{fig:factpose} before rendering.  As Figure~\ref{fig:car_2d} shows, we
generate good reconstructions and point clouds.

To quantify the quality of our reconstructions, we compare the PSNR, SSIM, LPIPS~\cite{DBLP:conf/cvpr/ZhangIESW18}
with supervised NeRF (using pre-processed pose information), and NeRF{-}{-}.  As Table~\ref{tab:quant:2d} shows,
for face-forwarding scenes, our method achieves reasonably high performance.  We cannot beat NeRF{-}{-} because we
aim to solve a more general pose prediction problem from a single input, instead of fitting camera parameters as
trainable variables.
For $360^\circ$ scene views, which NeRF{-}{-} completely fails to handle, our method still yields good
reconstructions.  This validates the design choice of our multi-pose rendering system in tolerating large camera pose
perturbations.

\subsection{Visualization of Pose Optimization}
\begin{figure*}[t]
   \begin{center}
   \begin{minipage}[t]{0.16\linewidth}
      \vspace{0pt}
      \centering
      \subfigure[\textbf{\textsf{\scriptsize{Fern GT.}}}]{
        \includegraphics[width=\linewidth]{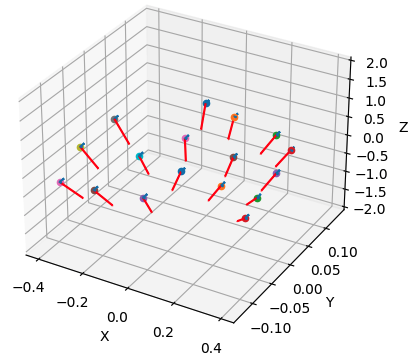}
      }
   \end{minipage}%
   \begin{minipage}[t]{0.16\linewidth}
      \vspace{0pt}
      \centering
      \subfigure[\textbf{\textsf{\scriptsize{Fern Pred.}}}]{
         \includegraphics[width=\linewidth]{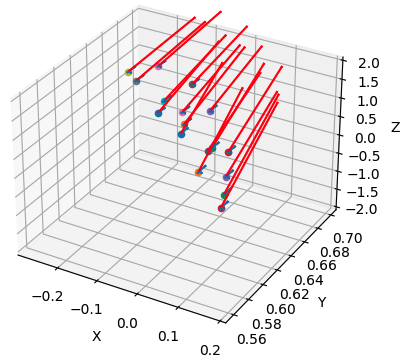}
      }
   \end{minipage}
      \begin{minipage}[t]{0.16\linewidth}
      \vspace{0pt}
      \centering
      \subfigure[\textbf{\textsf{\scriptsize{Car GT.}}}]{
         \includegraphics[width=\linewidth]{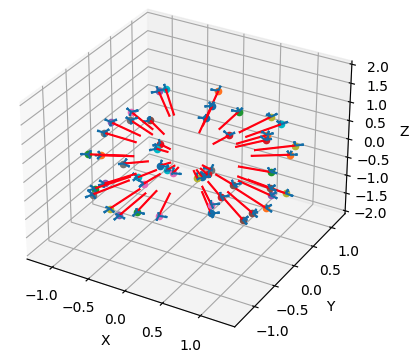}
      }
   \end{minipage}
    \begin{minipage}[t]{0.16\linewidth}
      \vspace{0pt}
      \centering
      \subfigure[\textbf{\textsf{\scriptsize{Car Pred.}}}]{
         \includegraphics[width=\linewidth]{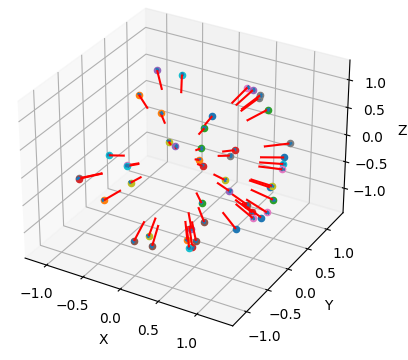}
      }
   \end{minipage}
   \begin{minipage}[t]{0.16\linewidth}
      \vspace{0pt}
      \centering
      \subfigure[\textbf{\textsf{\scriptsize{Lego GT.}}}]{
        \includegraphics[width=\linewidth]{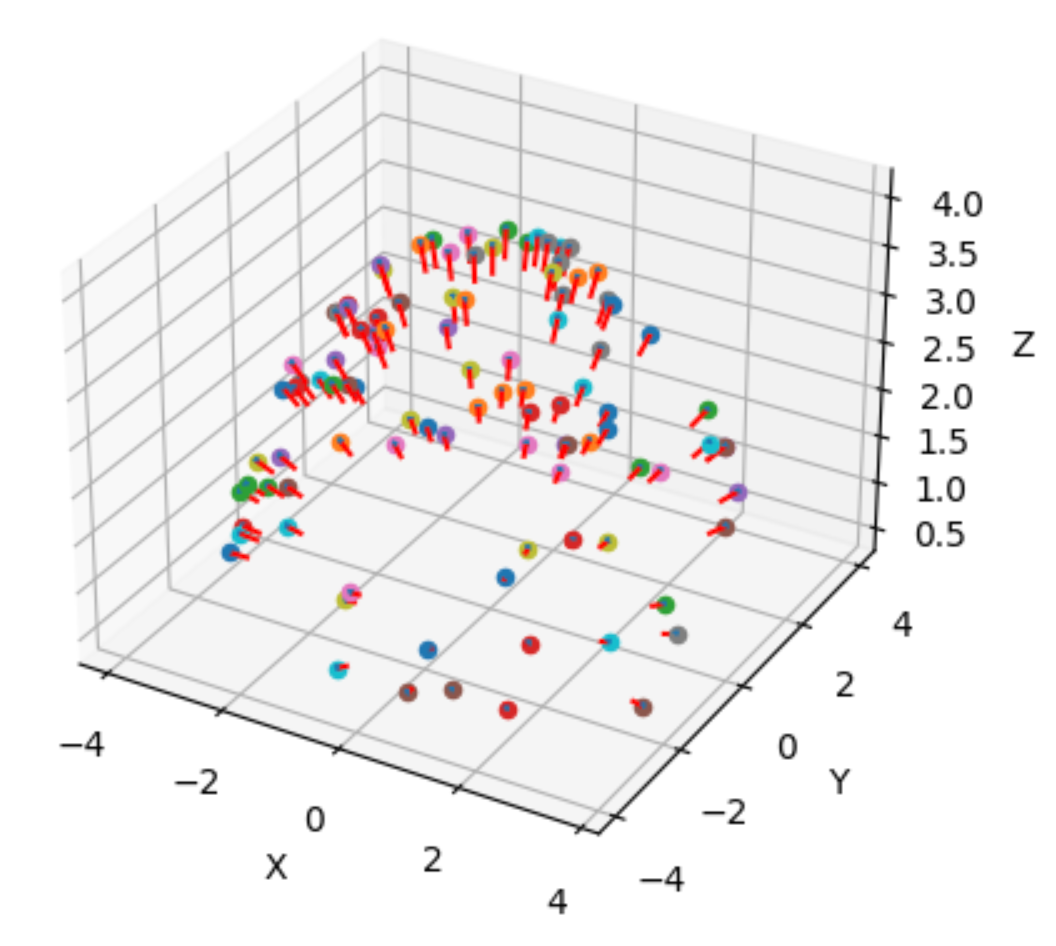}
      }
   \end{minipage}%
   \begin{minipage}[t]{0.16\linewidth}
      \vspace{0pt}
      \centering
      \subfigure[\textbf{\textsf{\scriptsize{Lego Pred.}}}]{
        \includegraphics[width=\linewidth]{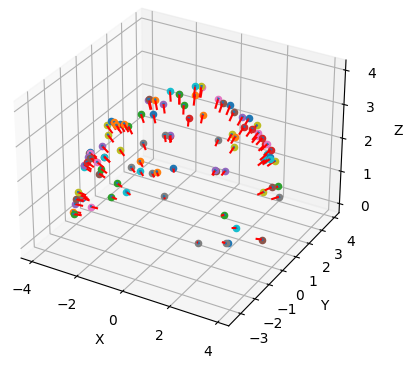}
      }
   \end{minipage}%
   \end{center}%
   \vspace{-1em}
   \caption{%
      \textbf{Visualization of camera poses for Fern, Car, and Lego.}
   }%
   \label{fig:camera_pred}
\end{figure*}

As shown in Figure~\ref{fig:camera_pred}, we also demonstrate that our pose prediction system can generate reasonable
pose estimates, though not in the same coordinate system, compared with ground-truth cameras.

\begin{figure*}[t!]
   \begin{minipage}[tbh]{0.7\linewidth}
   \includegraphics[width=\linewidth]{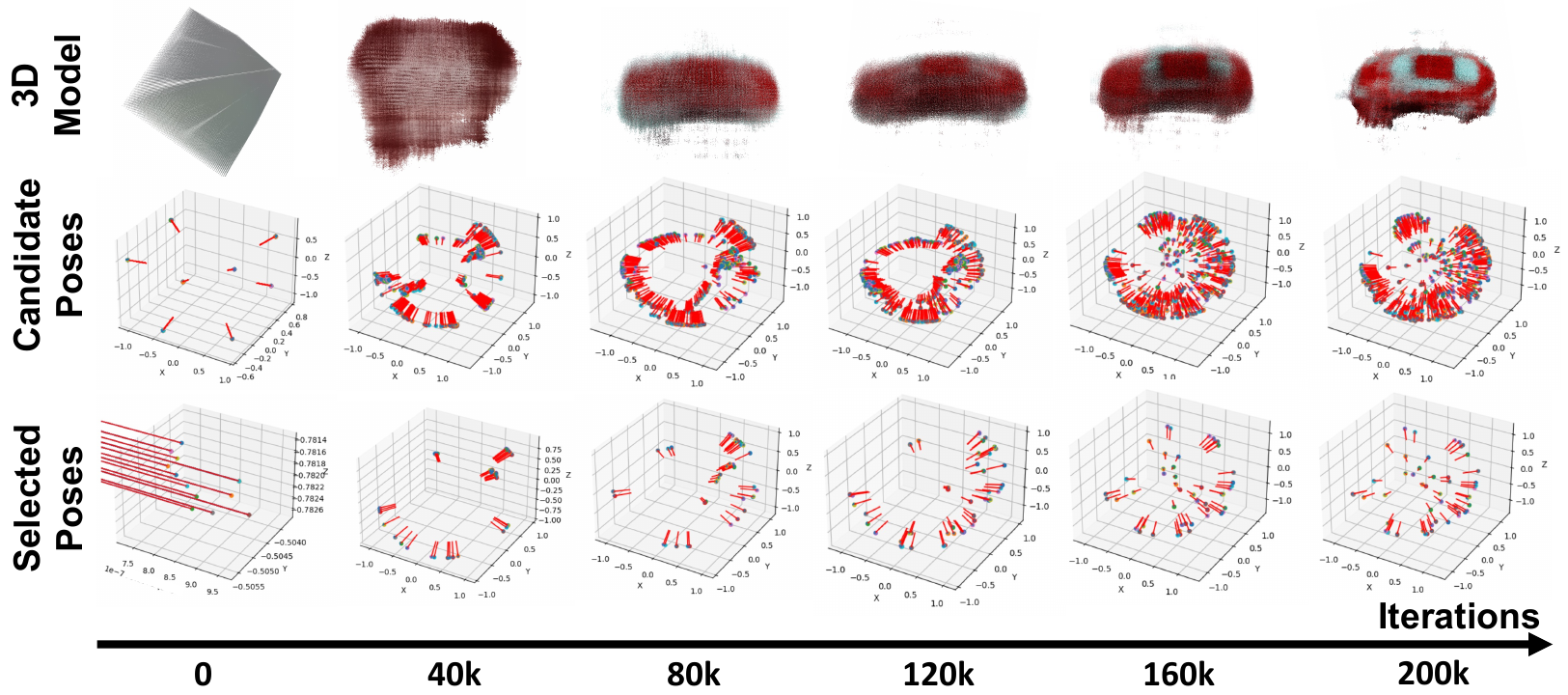}
   \vspace{-1.5em}
   \caption{%
      \textbf{{Joint 3D and pose optimization for Car during training.}
   }}%
   \label{fig:camera_evo}
   \end{minipage}
   \begin{minipage}[tbh]{0.28\linewidth}
      \footnotesize
      \centering
      \begin{tabular}{lccccccccc}
         \toprule
         &\multicolumn{1}{c}{{$\lambda$}}
         &\multicolumn{1}{c}{Acc.}
         &\multicolumn{1}{c}{PSNR} \\
         \midrule
         &0.01 &28 &\xmark \\
         &0.02 &72 & 19.43 \\
         &0.05 &66 & 18.88\\
         &0.1 &98 &26.43\\
         &0.2 &90 &22.19\\
         &0.5 &100 &\xmark \\
         &1.0 &100 &\xmark \\
         \bottomrule
      \end{tabular}
      \vspace{-3pt}
      \captionof{table}{\textbf{Effect of $\lambda$.}}
      \label{tab:lambda}
   \end{minipage}
\end{figure*}

\noindent\textbf{Camera distribution evolution during optimization}
To better demonstrate the pose prediction refinement during the optimization process, we visualize the camera poses at
different training iterations for the Car scene.  As shown in Figure~\ref{fig:camera_evo}, the candidate poses refers
to all possible predictions over 8 quaternions while the selected poses represent those with maximum classification
scores.  During training, the learned candidate poses tend to cover the sphere uniformly, with the selected pose
distributions gradually converge to that provided by the pre-processed dataset.  We observe simultaneous refinements
of both 3D model and pose prediction along the training process.

\subsection{Novel View Synthesis}
\begin{figure*}[t]
   \begin{minipage}[b]{0.481\linewidth}
      \vspace{0pt}
      \subfigure[\textbf{\textsf{\scriptsize{Car3D.}}}]{
         \includegraphics[width=\linewidth]{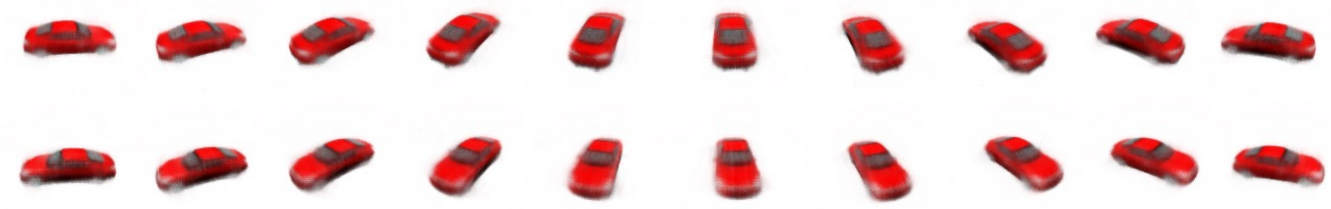}
      }
      \subfigure[\textbf{\textsf{\scriptsize{Lego.}}}]{
         \includegraphics[width=\linewidth]{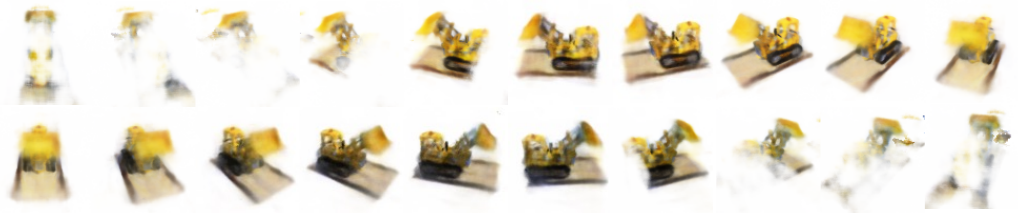}
      }
      \subfigure[\textbf{\textsf{\scriptsize{Drums.}}}]{
         \includegraphics[width=\linewidth]{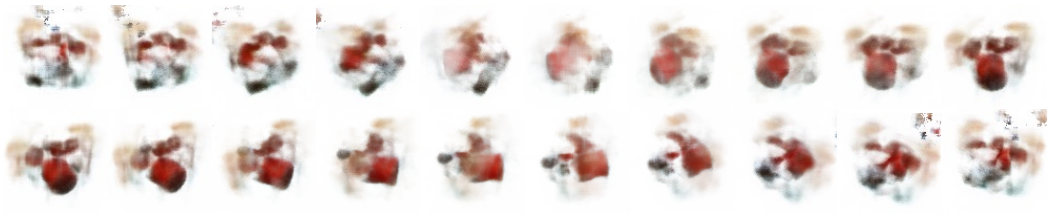}
      }
      \vspace{-10pt}
      \caption{\textbf{\textsf{\scriptsize{{Novel view synthesis from circle trajectory.}}}}}%
      \label{fig:gen_car_camera1}
   \end{minipage}
   \begin{minipage}[b]{0.48\linewidth}
      \vspace{0pt}
      \includegraphics[width=\linewidth]{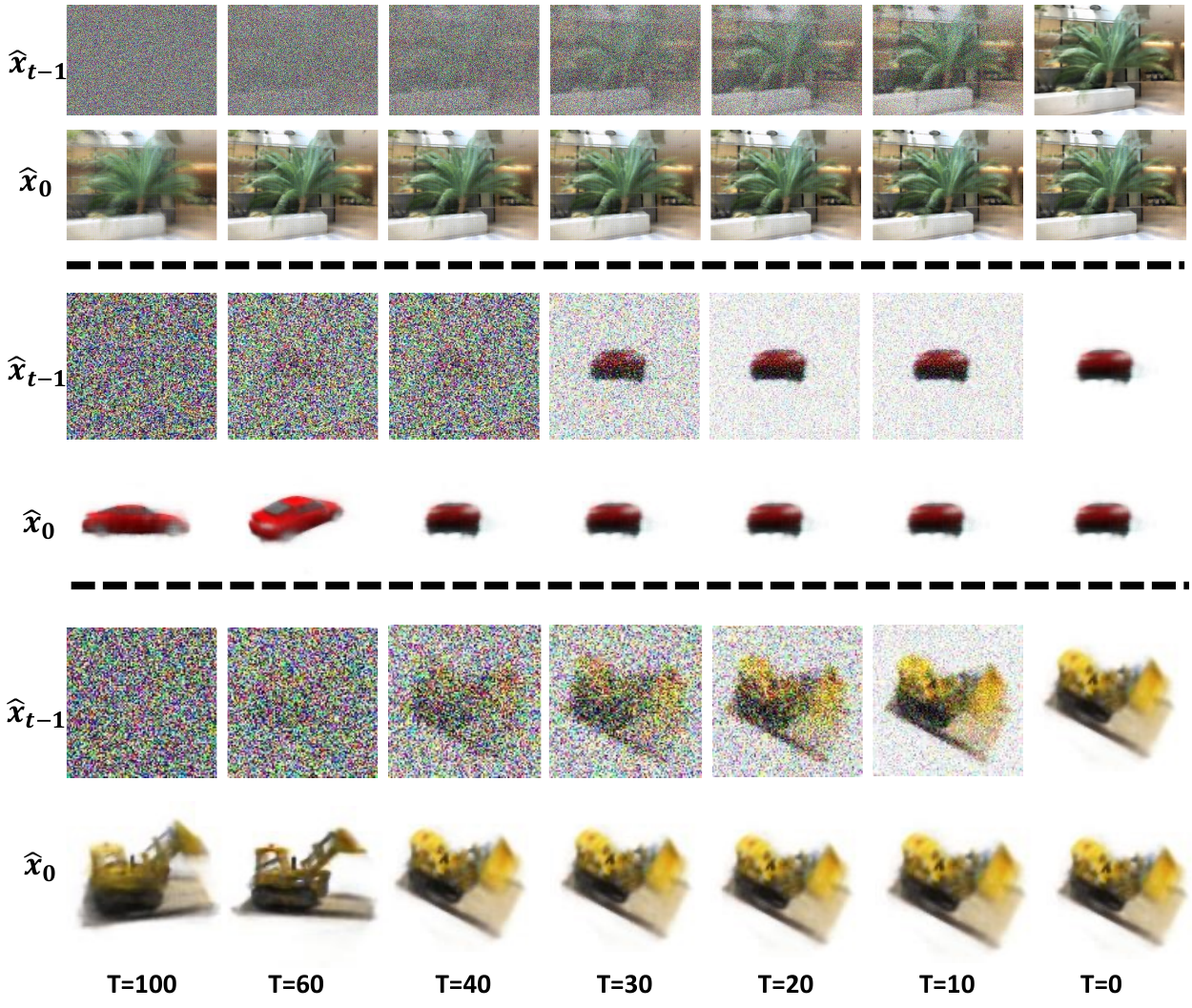}
      \vspace{-8pt}
      \caption{\textbf{\textsf{\scriptsize{{Novel view synthesis from Gaussian noise.}}}}}%
   \label{fig:gen_fern_noise}
   \end{minipage}
\end{figure*}

\textbf{From camera trajectory:}
We show that our joint-learned 3D model, even learned with unknown pose, can generate novel views using a manually
designed camera trajectory as in a supervised NeRF.  In Figure~\ref{fig:gen_car_camera1}, we generate novel views
using a continuous spiral camera path (pointing to the origin) for three different challenging scenes.

\textbf{From Gaussian noise:}
Given the nice property of denoising diffusion training, we have the flexibility to generate novel views from Gaussian
noise progressively.  In the DDPM Markov process, the model implicitly formulates a mapping between noise and data
distributions.  We validate this by visualizing the reconstructed $\hat{x}_{t-1}$ and $\hat{x}_{0}$ along the
sequential reversed diffusion process in Figure~\ref{fig:gen_fern_noise}.  We observe gradual refinement as denoising
steps approach $t=0$.

\begin{figure*}[t]
   \vspace{-1em}
   \begin{minipage}[b]{0.48\linewidth}
      \vspace{0pt}
      \subfigure[\textbf{\textsf{\scriptsize{Ours}}}]{
         \includegraphics[width=0.2\linewidth]{./figs/car0.png}
      }
      \vspace{0pt}
      \subfigure[\textbf{\textsf{\scriptsize{AE}}}]{
         \includegraphics[width=0.2\linewidth]{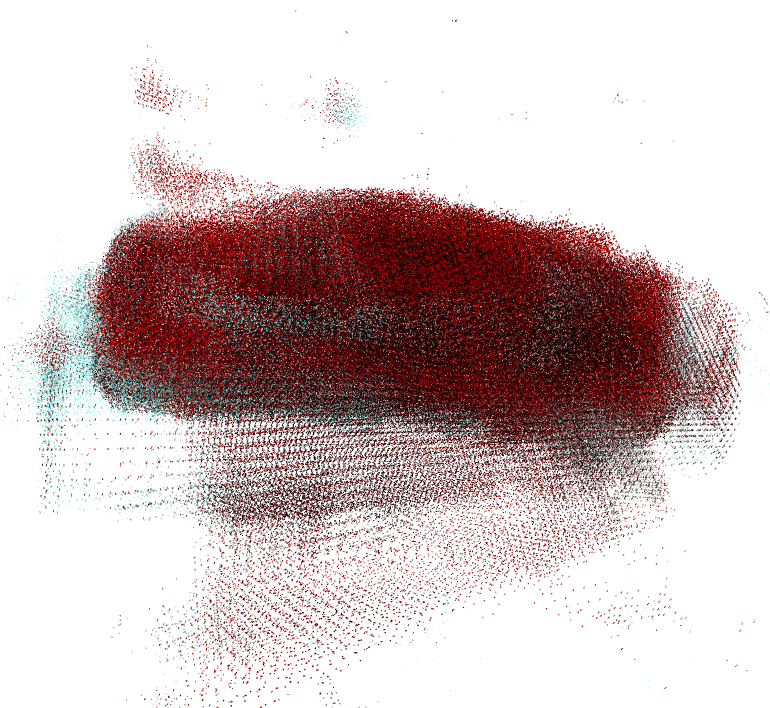}
      }
      \vspace{0pt}
      \subfigure[\textbf{\textsf{\scriptsize{AE-Pred}}}]{
         \includegraphics[width=0.2\linewidth]{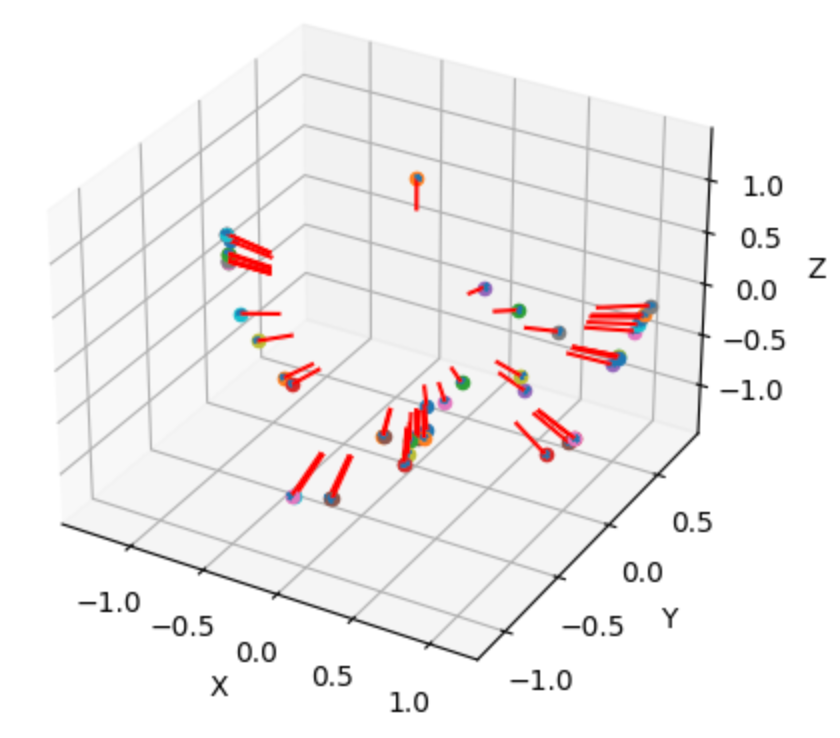}
      }
      \vspace{0pt}
      \subfigure[\textbf{\textsf{\scriptsize{GT}}}]{
         \includegraphics[width=0.2\linewidth]{./figs/gt_cam_car.png}
      }
      \caption{\textbf{\textsf{\scriptsize{{Learning w/o denoising diffusion.}}}}}%
      \label{fig:ablation:clean}
   \end{minipage}
   \begin{minipage}[b]{0.48\linewidth}
      \vspace{0pt}
      \subfigure[\textbf{\textsf{\scriptsize{4$\times$}}}]{
         \includegraphics[width=0.21\linewidth]{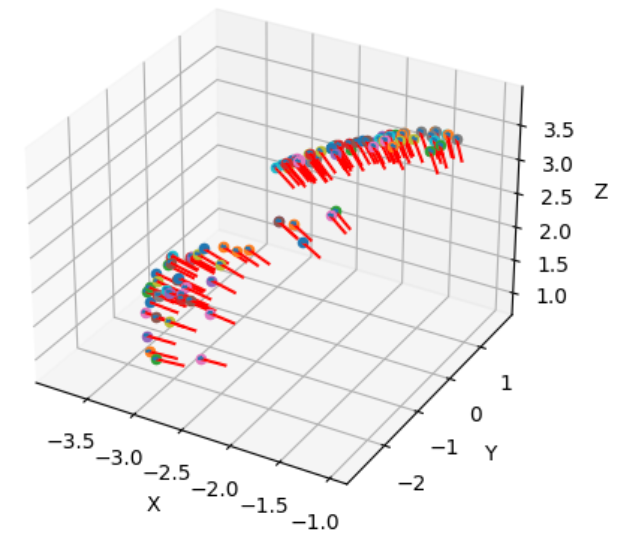}
      }
      \vspace{0pt}
      \subfigure[\textbf{\textsf{\scriptsize{8$\times$}}}]{
         \includegraphics[width=0.21\linewidth]{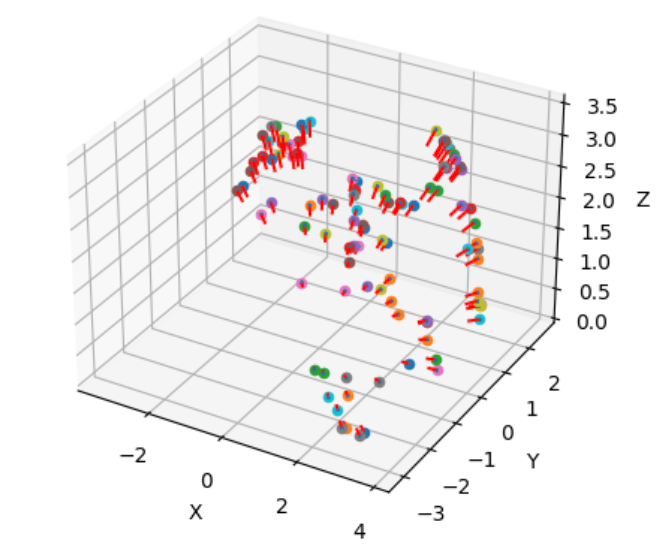}
      }
      \vspace{0pt}
      \subfigure[\textbf{\textsf{\scriptsize{12$\times$}}}]{
         \includegraphics[width=0.21\linewidth]{./figs/predicted_cam_lego.png}
      }
      \vspace{0pt}
      \subfigure[\textbf{\textsf{\scriptsize{GT}}}]{
         \includegraphics[width=0.21\linewidth]{./figs/gt_cam_lego.png}
      }
      \caption{\textbf{\textsf{\scriptsize{{Poses learned with different camera number.}}}}}%
      \label{fig:ablation:cam_num}
   \end{minipage}
\end{figure*}

\subsection{Abalation Studies}
\noindent\textbf{Training with clean images.}
Training NeRF with denoising diffusion is an essential part of our method.  We show the effectiveness of this design
by varying the input with clean images in our method, in which case our architecture downgrades to an autoencoder (AE).  As
shown in Figure~\ref{fig:ablation:clean}, the baseline trained with clean images (denoted as AE) yields an incorrect 3D
model for the Car scene.  This suggests the failure of camera pose prediction, hence leading to a NeRF over-fitting
issue.  Moreover, AE fails to perform novel-view synthesis given a pre-defined camera trajectory, as shown in
Figure~\ref{fig:gen_car_AE}.  Thus, denoising diffusion training not only provides a new way to perform novel view
synthesis, but also significantly improves the learned 3D reconstruction.

\noindent\textbf{Candidate camera numbers in multi-pose rendering.}
For Lego and Drums on a semi-sphere, we choose to use 12 candidate cameras instead of 4 at initialization for each
view.  The 4-way case poses an easier camera classification task to the system, but restricts flexibility for
discovering view correspondence and thereby pushes the NeRF to overfit on incorrect pose predictions.  Increasing
capacity to 12 cameras during training addresses this issue and prevents the system from converging to a suboptimal
solution.  As shown in Figure~\ref{fig:ablation:cam_num}, $4\times$ and $8\times$ variants fail to converge to the
correct pose distributions, while the $12\times$ succeeds.

\noindent\textbf{Trade-off between classification and reconstruction.}
Due to the discrepancy between training and testing in our multi-pose system, the quality of rendering highly relies on
the camera classification accuracy.  To evaluate the performance of this self-supervised classifier, we use the camera
index which produces the minimum reconstruction loss as ground-truth.  We study the effect of the classification loss
term by alternating $\lambda\in\{0.01, 0.02, 0.05,0.1,0.2, 0.5,1.0\}$, generating trade-offs between the accuracy
and PSNR for Car.  As shown in Table~\ref{tab:lambda}, NeRF training yields a poor reconstruction and even an
optimization failure when the classification accuracy is low.  When we set $\lambda$ as a large value, the model runs
into a local minimum: the classifier obtains $100\%$ accuracy at early training iterations and poses cannot be
jointly optimized to discover correspondence.

\section{Conclusion}
We propose a novel technique that places NeRF inside a probabilistic diffusion framework to accurately predict
camera poses and create detailed 3D scene reconstructions from collections of 2D images.  Our approach enables
training NeRF from images with unknown pose.  Using a carefully constrained architecture and differentiable volume
renderer, we learn a camera pose predictor and 3D representation jointly.  Our experimental results and ablation
studies confirm the effectiveness of this method, demonstrating its capability to produce high-quality reconstructions,
localize previously unseen images, and sample novel-view images, all while trained in an entirely unsupervised manner.

\small
\bibliography{ref}
\bibliographystyle{plainnat}

\appendix
\newpage
\section{Appendix}
\begin{figure*}[tbh]
   \begin{center}
    \includegraphics[width=\linewidth]{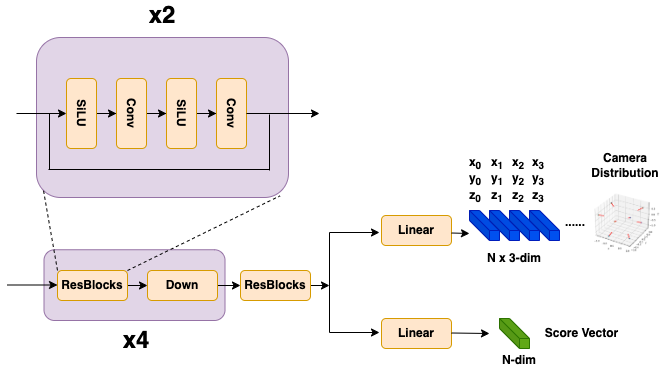}
   \end{center}
   \caption{%
      \textbf{Detailed architecture of pose prediction network.}
      A shared encoder trunk processes an input image and branches into heads for predicting a set of candidate
      camera poses, as well as a score vector indicating the probability the image was acquired from each of the
      predicted cameras.%
   }%
   \label{fig:3darch_detail}
\end{figure*}

\renewcommand{\lstlistingname}{Code}
\begin{lstlisting}[language=Python, label={lst:transform_xyz}, caption=Obtain rotation matrix from camera position]
import torch as th
import pytorch3d.transforms as tf
def gen_rotation_matrix_from_xyz(xyz, in_plane=th.from_numpy(np.array([0.0, 0.0, 0.0])).cuda().float()):
    cam_from = xyz
    cam_to = th.from_numpy(np.zeros(3)).float().cuda()
    tmp = th.from_numpy(np.array([0.0, 1.0, 0.0])).float().cuda()

    diff = cam_from - cam_to
    forward = diff / th.linalg.norm(diff)
    crossed = th.cross(tmp, forward)
    right = crossed / th.linalg.norm(crossed)
    up = th.cross(forward, right)

    R = th.stack([right, up, forward])
    R_in_plane = tf.rotation_conversions.euler_angles_to_matrix(in_plane, "XYZ")
    return R_in_plane @ R
\end{lstlisting}

\begin{lstlisting}[language=Python, label={lst:lkat}, caption=Camera transformation with pointing to the origin.]
import torch as th
def lkat(eye, target, up):
    forward = normalize(target - eye)
    side = normalize(th.cross(forward, up))
    up = normalize(th.cross(side, forward))
    
    zero = th.zeros(1).float().cuda()
    one = th.ones(1).float().cuda()
    trans_v0 = th.cat([side[0:1], up[0:1], -forward[0:1], zero])  # (3, 1)
    trans_v1 = th.cat([ side[1:2], up[1:2], -forward[1:2],zero])
    trans_v2 = th.cat([side[2:3], up[2:3], -forward[2:3], zero])
    trans_v3 = th.cat([ -th.dot(side, eye)[None], -th.dot(up, eye)[None], th.dot(forward, eye)[None], one])
    c2w = th.stack([trans_v0, trans_v1, trans_v2,trans_v3], dim=0)
    return c2w
\end{lstlisting}

\begin{lstlisting}[language=Python, label={lst:pose_predict}, caption=Pose distribution prediction with 12$\times$ camera candidates]
import torch as th
eye_candidates = th.Tensor([[1,1,1],[1,-1,1],[-1,1,1],[-1,-1,1],
                            [1,1,1],[1,-1,1],[-1,1,1],[-1,-1,1],
                            [1,1,1],[1,-1,1],[-1,1,1],[-1,-1,1]]).cuda()
r = th.tensor([4.0]).float().cuda()
target = th.from_numpy(np.zeros(3)).float().cuda()
up = th.from_numpy(np.array([0.0, 1.0, 0.0])).float().cuda()

# h: the output of the last residual block in pose prediction model.
h1 = linear1(h.squeeze(-1)).squeeze() # (12*3, )
h2 = linear2(h.squeeze(-1)).squeeze() # (12, ), score vector
    
zero = th.zeros(1).float().cuda()
init_cam_pos = th.cat([ zero,    zero,   r)  # (3, 1)
all_poses = []

for index in range(12):
    h3= th.sigmoid(h1[3*index:index*3+3])
    h3 = th.diag(eye_candidates[index])@h3
    h3 = h3/th.linalg.norm(h3)
    R1 = gen_rotation_matrix_from_xyz((h3)
    eye = (R1 @ init_cam_pos)
    look_at1 = lkat(eye, target, up)
    pose1 = th.eye(4).float().cuda()
    pose1[:3, :3] = look_at1[:3, :3]
    pose1[:3, 3] = -look_at1[:3, :3] @ look_at1[3, :3]
    all_poses.append(pose1[:3,:4])
all_poses = th.stack(all_poses, 0)
return all_poses, h2 #pose distribution (12,3,4), scores (12, 1)
\end{lstlisting}

\begin{figure*}[tbh]
\subfigure[\textbf{\textsf{\scriptsize{Ours}}}]{
\includegraphics[width=\linewidth]{./figs/car_gen_camera1.pdf}
}
\subfigure[\textbf{\textsf{\scriptsize{AE Baseline}}}]{
\includegraphics[width=\linewidth]{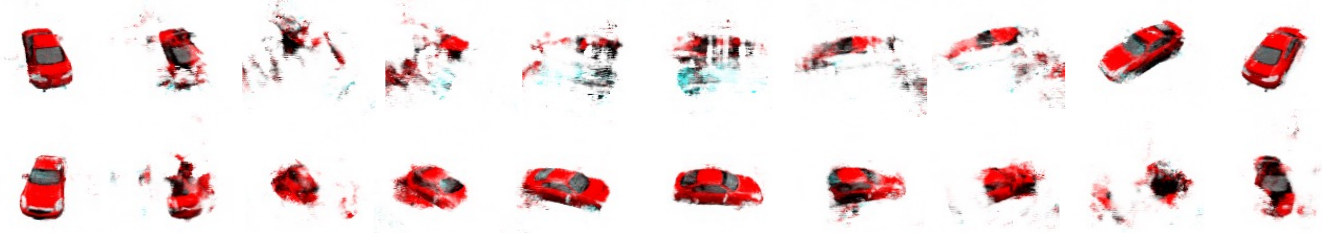}}
    \caption{%
      \textbf{Diffusion training benefits novel view synthesis on Car3D.}
      Our system, wrapped within a DDPM for training, significantly outperforms the same architecture trained as
      a simple autoencoder (AE).  Training with the more challenging denoising task yields more robust generalization
      for the pose prediction network and NeRF scene representation.%
   }%
   \label{fig:gen_car_AE}
\end{figure*}


\end{document}